\documentclass{article}
\usepackage{PRIMEarxiv}

\usepackage{graphicx}
\usepackage{amsmath}
\usepackage{amssymb}
\usepackage{booktabs}
\usepackage{multicol}
\usepackage{multirow}
\usepackage{caption}
\usepackage{floatrow}
\usepackage[dvipsnames]{xcolor}
\usepackage{soul}

\usepackage[pagebackref,breaklinks,colorlinks]{hyperref}

\usepackage[capitalize]{cleveref}
\crefname{section}{Sec.}{Secs.}
\Crefname{section}{Section}{Sections}
\Crefname{table}{Table}{Tables}
\crefname{table}{Tab.}{Tabs.}

\def\cvprPaperID{9735} 
\def\confName{CVPR}
\def\confYear{2022}

\newcommand{\toremove}[1]{{\color{green} [{\bf remove}: #1}]}

\definecolor{ForestGreen}{RGB}{34,139,34}
\newcommand{\wbc}[1]{\textcolor{ForestGreen}{wbc: #1}}

\newif\ifclean
\cleantrue

\ifclean
\renewcommand{\toremove}[1]{}
\fi

\begin{document}

\title{A Fistful of Words: Learning Transferable Visual Models\\
from Bag-of-Words Supervision}

\author{
  Ajinkya Tejankar\thanks{The work was mainly done during internship
at Facebook AI.} \\
  University of Maryland, Baltimore County \\
  \texttt{at6@umbc.edu} \\
  \And
  Maziar Sanjabi \\
  Facebook AI \\
  \texttt{maziars@fb.com} \\
  \And
  Bichen Wu \\
  Facebook AI \\
  \texttt{wbc@fb.com} \\
  \And
  Saining Xie \\
  Facebook AI \\
  \texttt{s9xie@fb.com} \\
  \And
  Madian Khabsa \\
  Facebook AI \\
  \texttt{mkhabsa@fb.com} \\
  \And 
  Hamed Pirsiavash \\
  UC Davis \\
  \texttt{hpirsiav@ucdavis.edu} \\
  \And
  Hamed Firooz \\
  Facebook AI \\
  \texttt{mhfirooz@fb.com} \\
}
\maketitle

\begin{abstract}

Using natural language as a supervision for training visual recognition models holds great promise. Recent works have shown that if such supervision is used in the form of alignment between images and captions in large training datasets, then the resulting aligned models perform well on zero-shot classification as downstream tasks\footnote{In this paper, zero-shot refers to the case where no labeled data is provided for downstream tasks while the category names are known.}. In this paper, we focus on teasing out what parts of the language supervision are essential for training zero-shot image classification models. Through extensive and careful experiments, we show that: 1) A simple Bag-of-Words (BoW) caption could be used as a replacement for most of the image captions in the dataset. Surprisingly, we observe that this approach improves the zero-shot classification performance when combined with word balancing. 2) Using a BoW pretrained model, we can obtain more training data by generating pseudo-BoW captions on images that do not have a caption. Models trained on images with real and pseudo-BoW captions achieve stronger zero-shot performance. On ImageNet-1k zero-shot evaluation, our best model, that uses only 3M image-caption pairs, performs on-par with a CLIP model trained on 15M image-caption pairs (31.5\% vs 31.3\%).

\end{abstract}



\section{Introduction}

\begin{figure}
    \centering
    \includegraphics[width=.7\columnwidth]{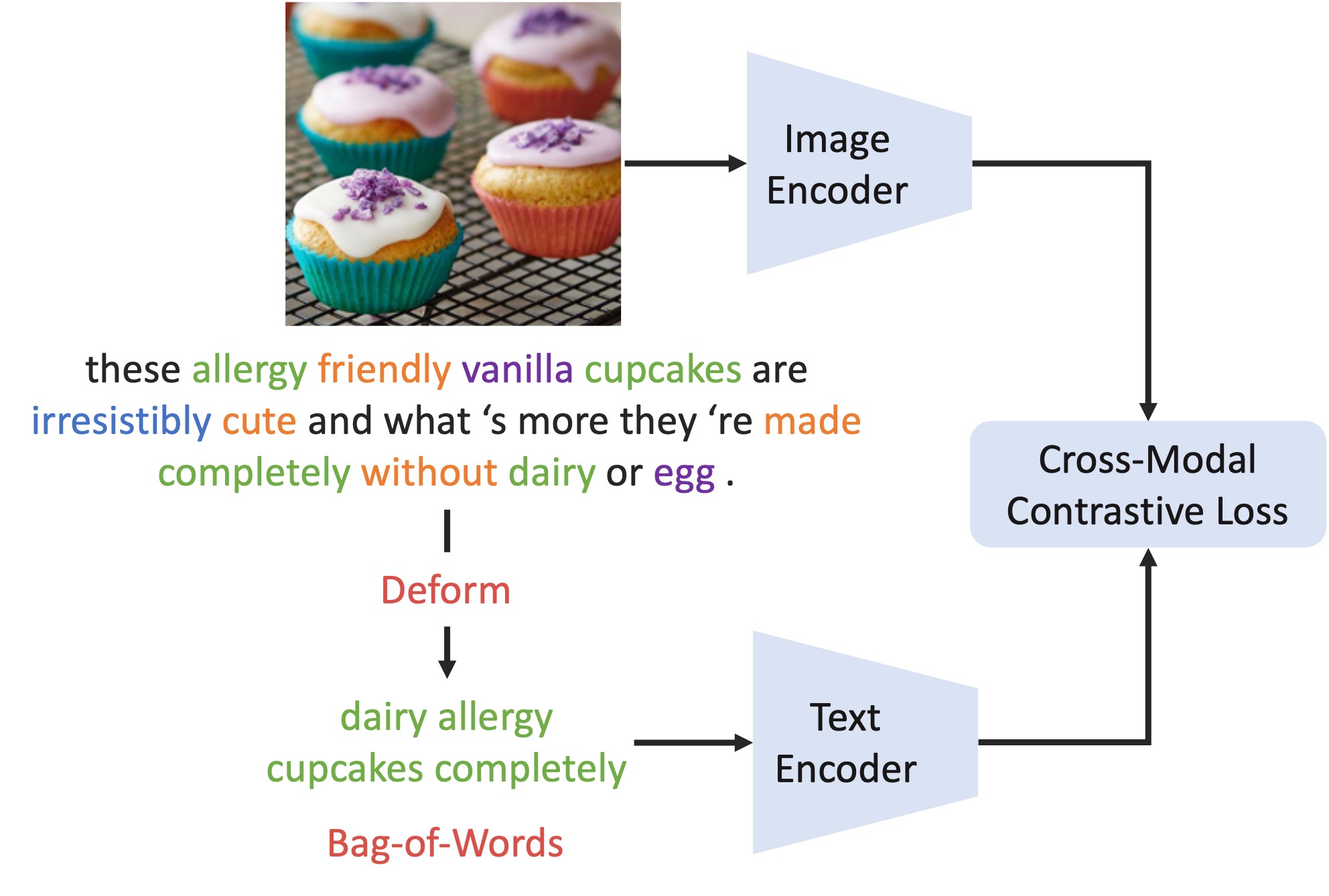}
    \caption{\textbf{Overview of our method:} We train image and text encoders with a cross-modal contrastive loss, similar to \cite{convirt,clip}, with image and caption pairs, but we design various operators to deform the intact caption. For example, shuffling the words, removing certain words, etc. Colors except green, represents words removed by different deformation operators. The resulting caption is a Bag-of-Words with only a few words from the original caption. We show that a model trained with BoW captions performs on-par or better than the model trained with intact captions (29.5\% vs 30.1\% on ImageNet-1k zero-shot).}
    \label{fig:teaser}
\end{figure}

\label{sec:intro}
Recently, CLIP \cite{clip} showed impressive results on zero-shot learning, demonstrating the ability to generalize to unseen image recognition tasks. CLIP achieves this by first training vision and text encoders to distinguish correct image-captions pairs, and then using the text encoder to convert category names to embedding vectors, which serve as a classifier's weights. 
A key ingredient for the success of CLIP's \cite{clip} zero-shot performance is to use a large-scale (400 million) dataset with paired images and captions to train the vision models. Compared with the traditional supervised learning datasets that require human annotations, image-text data are easier to obtain. Moreover, captions also contain much richer information about images than categorical labels. Therefore, they can provide better supervision that leads to improved generalization.

However, information encoded in captions may be redundant or irrelevant to their paired images. For example, though many words in an image caption are important for the syntactical and grammatical properties of a sentence, they might not contribute to the semantics of the sentence and may be irrelevant to the content of the paired image. A natural question is, which parts of the captions are necessary to train vision models, or more specifically, zero-shot classification models? Can we find a simpler supervision to replace text captions in training vision models? Studying this problem is meaningful both scientifically and practically. It can help shed light on how humans might connect language and vision concepts, and also help us design more efficient learning algorithms to train computer vision models.




In this work, we investigate and tease out which parts of caption supervision is necessary for training vision models for zero-shot classification. To be more specific, we ask: 1) Do we really need natural language captions to train vision models?
2) If we can find a simpler supervision than text captions, can we use it to train vision models more efficiently by leveraging large volume of images that are not aligned with text captions?

To answer the first question, we conduct experiments by deforming the text captions, changing their language properties, and observe the impact on the trained model's zero-shot performance (see Figure \ref{fig:teaser} for an illustration of this process). For example, we use the word shuffling operations to break a sentence to a Bag-of-Words (BoW), we drop words in a sentence based on their positions or according to a limited vocabulary, etc. We observe that removing syntactic information in the captions and converting them into BoW does not degrade the trained model's zero-shot performance. Surprisingly, we can even improve the accuracy of the model trained on BoW captions by using a simple strategy of balancing the frequency of words. An explanation for the model's robustness to caption deformations can be found in the way we initialize the text encoder. In keeping with the broader theme of this work of relying less on supervision, we use unimodal self-supervised models like BYOL \cite{byol} and BERT \cite{bert} for initializing vision and text encoders respectively. Several studies in natural language understanding domain have shown that models based on BERT \cite{bert} can learn good features despite input words being shuffled during inference \cite{unnatural} or even during pre-training \cite{mlm_shuffle}. These findings suggest that deep contextual embedding models don't rely on language syntax for learning good representations.

For the second question, given the observation that BoW can provide supervision that is at least on-par with their original captions, we provide a simple recipe to further boost the zero-shot performance of vision models by leveraging images that are unaligned (without associated captions). Inspired by self-training framework \cite{self-training}, we use the model pre-trained on image-caption pairs to find nearest neighbors of a given unaligned image within a set of intact captions. Then, by ranking and filtering the words of those retrieved captions we generate a BoW caption for the unaligned image. Finally, we train a new model on the union of the original dataset and the new dataset with pseudo captions. Since only a subset of the data is aligned, we refer to this setting as ``semi-aligned'' learning similar to semi-supervised learning.

In summary, we discover that image captions with intact natural language structure are redundant for training vision models for zero-shot classification. In fact, deforming most of the text captions and turning them into BoW can even lead to improvements (29.5\% vs. 30.1\% on ImageNet-1k zero-shot). Based on this insight, we propose a method to generate pseudo BoW labels for images without captions. We train a model with 3M image-caption pairs and 3M unaligned images and further boost the zero-shot performance (30.1\% to 31.5\% on ImageNet-1k zero-shot). This model is on-par with CLIP trained on 15M image-caption pairs (31.5\% vs 31.3\% in ImageNet-1k zero-shot).

\section{Method}

\begin{figure*}
    \centering
    \includegraphics[width=\textwidth]{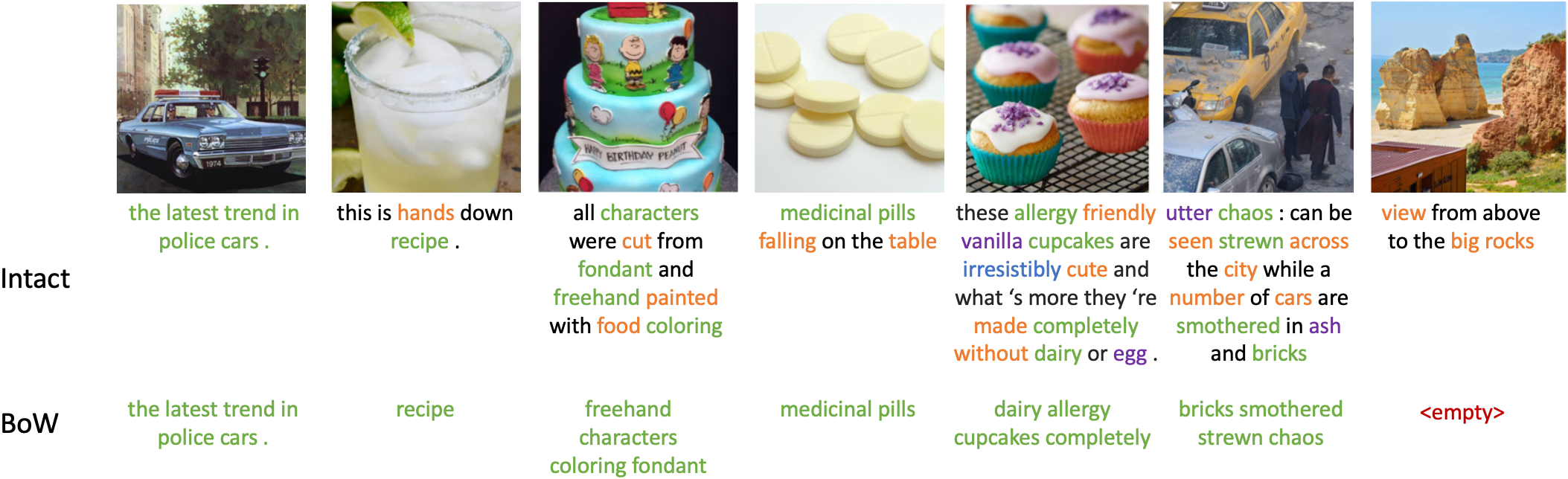}
    \caption{\textbf{From intact captions to BoW.} Above figure illustrates the process of converting intact captions to BoW. Notice that BoW are much denser with very little distracting information. The words are first shuffled to remove syntax by \texttt{Shuffle}. Next, while stop words (``all", ``were", ``the", etc.) and non-alphabetical tokens (``:" and ``.") are important parts of the syntax of sentences, none of them describe anything in their corresponding images. Hence, they are removed with \texttt{RmStopNalpha}. Next, words like ``cars", ``table", ``men", "rocks" etc. (orange color) are important for describing the image, but because they already appear frequently we can remove them with \texttt{RmTopFreq(1000)}. Note that the model still sees and learns about them but from the small set of captions that are kept intact (base set). Further, the blue word ``mikoshi" is a rare word that does not appear in the base set vocabulary, and hence maybe removed without hurting the model. It is removed by \texttt{LimitToBaseVocab}. Purple words like ``utter" and ``ash" are removed by \texttt{Keep(n)} which may actually remove useful words, yet BoW improve the performance. Finally, some captions will become empty since none of the words in them describe anything new. Thus, they are removed from the dataset.}
    \label{fig:visualize-intact-vs_defm}
\end{figure*}

The primary question explored in this work is: what parts of language supervision are necessary for training vision models for zero-shot (ZS) classification? Since our exploration reveals that intact captions are equivalent to their BoW couterparts, we use this insight to further boost the zero-shot performance by leveraging additional unaligned images (without captions). We first introduce the cross-modal contrastive pre-training framework used in all of our experiments. Then, we introduce various techniques of deforming the text, thereby converting the captions to BoW. Finally, we describe the process of assigning BoW pseudo-labels to additional unaligned images.

\subsection{Cross-modal Contrastive Pre-training}

We need to align the features of images and text before they can be used for zero-shot learning. The alignment is done by training the model to distinguish between the correct image-caption pairs from incorrect ones. We first initialize both image and text encoders with self-supervised learning (SSL) models trained in respective modalities, and then align their representations by training them on a dataset of image-caption pairs.

Given a pair of image and caption, denoted as $(x_i, y_i)$, we use image encoder $f$ and text encoder $g$, to extract a pair of feature embeddings as $(u_i, v_i)$, where $u_i = f(x_i), v_i = g(y_i)$. Similar to \cite{clip} and \cite{convirt}, we use a cross-modal InfoNCE \cite{infonce} loss. In order to increase the number of negative samples without increasing the batch size and training memory, we follow \cite{moco} and maintain a memory bank of negative samples. Embedding vectors in the memory bank are computed by an exponential moving average (EMA) copy of the models, which we denote as $f', g'$. We also denote the EMA model's output as $u_i', v_i'$. Now we can write our training loss as:

\begin{align}
    L^{image}_i &= - \log \frac{\exp(d(u_i, {v'}_i)/\tau)}{\sum^K_{j=1} \exp(d(u_i, {v'}_j)/\tau)},  
\end{align}

\begin{align}
    L^{text}_i &= - \log \frac{\exp(d(v_i, {u'}_i)/\tau)}{\sum^K_{j=1} \exp(d(v_i, {u'}_j)/\tau)},
\end{align}

\begin{align}
    L &= \frac{1}{2B} \sum^B_{i=1} L^{image}_i + L^{text}_i,
\end{align}

Where, $B$ is batch size, $K$ is the size of the memory bank, $\tau$ is a temperature parameter, and $d(\cdot, \cdot)$ is the cosine similarity function.
We use the above loss function for all of our experiments.

\label{sec:text-deformation}
\subsection{From Captions to Bag-of-Words}

In order to understand the importance of natural language structure in captions, we design a set of operations to change/deform the captions. Each operation removes some syntactical or structural information from the original caption and converts it to BoW. If the lost information was used by model for zero-shot learning the model would degrade. We use following operations in our experiments.

\textbf{\texttt{Shuffle:}}~Shuffles the order of words in a given caption. The goal here is to remove syntactic information from text and convert them into BoW. While humans may have trouble understanding language without syntax, language models seem to be robust to text without syntax \cite{mlm_shuffle,unnatural}. 

\textbf{\texttt{RmStopNalpha:}}~Removes stop words and non-alphabetical words from a given caption, since removing them does not affect semantic content of a sentence. Non-alphabetical words tend to be numbers and punctuations which have little correspondence to the object present in the image. Similarly, stop words are needed in a sentence to make it syntactically correct but they do not correspond to objects in images like nouns or adjectives do. Since we only focus on English captions, we use the list of stop words from NLTK library \cite{nltk}.

\textbf{\texttt{LimitToBaseVocab:}}~Limits the words in a given caption to be from the vocabulary of a base set of captions. The goal is explore restricting the diversity of words in captions. Hence, we randomly select a small subset of all captions, 10\% in our experiments, to be the ``base'' captions. Then, we deform the 90\% split such that any word not present in the vocabulary of the base 10\% split is removed.

\textbf{\texttt{RmTopFreq(t):}}~Removes top-\texttt{t} most frequent words. We calculate the frequency of each word as the number of captions from the base set in which it appears. Also, similar to \texttt{LimitToBaseVocab}, this deformation is only applied on the 90\% set of captions. The goal is to balance the distribution of words by keeping the most frequent words only in the base 10\% split while removing them from 90\% of the captions.

\textbf{\texttt{Keep(n):}}~Keeps the first \texttt{n} words of a given caption in order to explore removing information from captions. This deformation is never used alone but always along with with other operations like \texttt{Shuffle} so that only a few random words from the original caption are retained. Note that except \texttt{Keep(n)}, the order in which other operations are applied does not matter. Hence, it is always applied in the end after all other operations.

These operations can be combined in different ways to increase or decrease the strength of deformation. The default deformation with highest strength that still works better than the intact captions baseline is the following cascade of operations: \texttt{Shuffle + RmStopNalpha + LimitToBaseVocab + RmTopFreq(1000) + Keep(4)}. Unless mentioned, this is the default deformation used for constructing BoW captions, and the resulting BoW dataset is referred to the ``default BoW" (see Figure \ref{fig:visualize-intact-vs_defm} for an illustration). Note that in some cases, captions may become empty if all their words are removed during deformation. In this case, the empty caption and the associated image is removed from the dataset which reduces the BoW dataset size. Finally, deformations are used to create a new BoW dataset which simply replaces the intact captions dataset during training.


\label{sec:semi-aligned}
\subsection{Utilizing Unaligned Images}

The insight that intact captions can be replaced with BoW without hurting the model's performance opens the opportunity to use unaligned images (images without associated captions) by pseudo-labeling them with BoW in a process similar to self-training \cite{self-training}. The goal is to leverage unaligned data which can be easier to acquire and maintain in comparison to aligned data. We call this setting semi-aligned learning since only a subset of data is aligned.

We first train a model with fully aligned set $S_a$ of image-caption pairs, and then use it to generate pseudo-labels for unaligned images. Given an unaligned image $x_i$, we augment it $c$ times with random cropping and resizing to get $c$ augmented images $\{x^b_i\}_{b=1:c}$. Next, we get features for each $x^b_i$ by passing it through the vision encoder $u^b_i = f(x^b_i)$. Then, for each $u^b_i$, we retrieve its top-$k$ nearest neighbors from a set of captions, and put them in a set $NN_i = \{y^j_i\}_{j=1:kc}$. The set of captions used for retrieval are intact and come from the set $S_a$ which was also used for training the fully-aligned model. Next, we use different strategies to rank the words in the $NN_i$ and then pick first $p$ words. The goal of these word ranking strategies is to find the set of words that best describe the image. See supplementary material for an illustration of this process and a few example pseudo labels.

\textbf{\texttt{MaxCount}} Rank each word according to the count of times it is present at least once in a caption $y_i^j \in NN_i$.
\textbf{\texttt{RmTopMaxCount}} Each word is ranked according to the \texttt{MaxCount} strategy, but similar to the text deformation discussed previously, 1000 most frequent words in the $S_a$ captions are removed. The goal is to focus more on the infrequent words. \textbf{\texttt{RmTopRand}} Rank the words randomly in \texttt{RmTopMaxCount} instead of ranking them based on count. \textbf{\texttt{WeightedCount}} Re-rank the words in \texttt{MaxCount} strategy by the inverse of their counts in the fully aligned caption set $S_a$. Specifically, rank each word $w_p \in NN_i$ by the ratio $count(w_p, NN_i)/count(w_p, S_a)$. Instead of removing the most frequent words this strategy simply ranks them lower.

\section{Experiments}

\begin{figure*}[t]
    \centering
    \includegraphics[width=\textwidth]{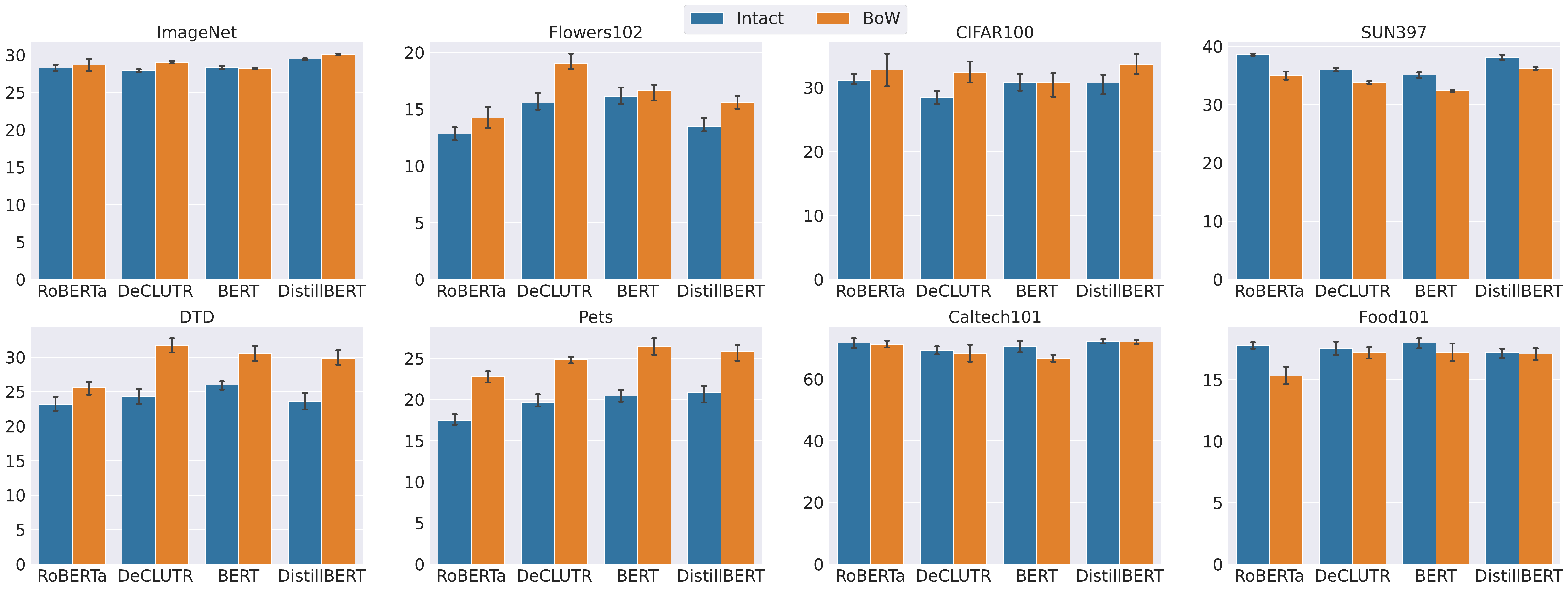}
    \caption{\textbf{Intact vs. BoW on additional datasets.} We compare intact and default BoW caption based models with 4 different architectures on zero-shot evaluation across 8 different datasets. The results show that for some datasets like, Flowers102, DTD, and Pets, all BoW models improve regardless of the architecture. While, for some of the datasets like, Caltech101, Sun397, and Food101, the accuracy of the deformed model consistently decreases. See the supplementary material for a detailed analysis of the results.}
    \label{fig:def-vs-no-def-all-datasets}
\end{figure*}

We use a ResNet-50 model trained with BYOL \cite{byol} on ImageNet \cite{imagenet} as the image encoder and transformer models trained with DeCLUTR (sci-base) \cite{declutr}, BERT (base-uncased) \cite{bert}, RoBERTa (base) \cite{roberta}, and DistilBERT (base-uncased) \cite{distilbert}. Note that although these models are trained on large unimodal datasets, their representations are not aligned across modalities. We use Conceptual Captions (CC) \cite{conceptual_captions} with ($\sim 2.9$M pairs) image-caption pairs as the source of aligned data, and 3M randomly sampled images from ImageNet-21k \cite{imagenet} as the source of unaligned data. We train for 20 epochs by default for CC-only experiments and 10 epochs for utilizing unaligned images (semi-aligned) experiments. Results for all experiments with 20 epochs are averaged over 4 runs with different seeds where BoW dataset is also constructed with a different seed. Further details can be found in the supplementary material.



\textbf{Zero-Shot Evaluation: } Primary zero-shot learning evaluation is done on ImageNet \cite{imagenet} dataset which has 1k categories. For the sake of brevity we use IN1k-ZS to refer to this evaluation. We also do zero-shot evaluation on a set of 7 fine-grained classification datasets: Food101 \cite{food101}, SUN397 \cite{sun397}, CIFAR100 \cite{cifar100}, Flowers102 \cite{flowers}, Pets \cite{pets}, Caltech-101 \cite{caltech101} and DTD \cite{dtd}. This is a subset of the datasets used in \cite{simclr,byol} for linear evaluation while CLIP \cite{clip} uses them for evaluating zero-shot learning. We follow the procedure in CLIP \cite{clip} for zero-shot learning. Given an image, we first forward it through the vision encoder to get its embedding, and find its nearest neighbor in the set of category embeddings. Category embeddings are created by first turning a category name into a sentence with a simple template \texttt{`a photo of a \{class name\}'} and then forward this sentence through the text encoder to get the category embedding. Unlike CLIP \cite{clip}, we don't use an ensemble of prompts for simplicity.



\subsection{Baselines}
First, we show that our training setup is competitive with other SOTA methods given similar number of aligned images. We compare our model trained for 40 epochs with intact captions and DistilBERT architecture. The comparison in Table \ref{tab:compare-other-methods} shows that our model trained on only $\sim$ 3M aligned data points is comparable to the CLIP \cite{clip} model trained on 15M aligned data points. Moreover, our results are higher compared to those of DeCLIP \cite{declip} and CLIP (implementation from \cite{declip}) on given same amount of aligned data. However, DeCLIP initializes their models from scratch while ours are initialized from SSL models. We believe this difference is the reason for the better performance of our models.

\begin{table}[h]
    \centering
    \begin{tabular}{lcc}
        \toprule
        Method & Dataset & IN1k-ZS \\
        \midrule
        CLIP \cite{clip} & YFCC-15M & 31.3 \\
        \midrule
        CLIP$^b$ & CC-3M & 20.6 \\
        DeCLIP$^a$ \cite{declip} & CC-3M & 27.2 \\
        Ours & CC-3M & 30.3 \\
        \bottomrule
    \end{tabular}
    \caption{\textbf{Comparison of our intact captions baseline with other methods}. We show that we use a strong intact captions baseline model for comparison with BoW models in following sections. Our model is better than DeCLIP and CLIP on CC-3M, and it is close to CLIP trained on YFCC-15M. $^a$ is a concurrent work under review. $^b$ refers to results of CLIP \cite{clip} implementation from \cite{declip}.}
    \label{tab:compare-other-methods}
\end{table}

\subsection{Intact Captions vs. Bag-of-Words}

In this section we experiment with deformations that convert intact captions to BoW. First, we compare the baseline trained with intact captions against BoW captions on IN-1k-ZS benchmark while accounting for convergence, variance due to randomness, and the effect of different text encoders. Next, we extend the comparison with two additional text encoders and 7 additional zero-shot (ZS) datasets. Then, we build intuition as to why BoW improves the performance when the words in them are balanced. Finally, we perform ablations over different deformations and hyperparameters. Please refer to the supplementary material for additional results on cross-modal retrieval.


\begin{table}
    \centering
    \begin{tabular}{lccll}
    \toprule
        \multirow{3}{*}{Epochs} & Avg. & $\Delta$ & \multicolumn{2}{c}{IN1k-ZS} \\
        & Cap & Dset &  DeCLUTR & DistilBERT \\
        & Len & Size &   & \\
        \midrule
        \multicolumn{4}{l}{\textit{intact captions}} \\
        20 & 10.3 & 0\% & 27.9 $\pm 0.2$ & 29.5 $\pm 0.1$ \\
        30 & 10.3 & 0\% & 28.2 & 29.8 \\
        40 & 10.3 & 0\% & 28.5 & 30.3 \\
        \midrule
        \multicolumn{4}{l}{\textit{default BoW}} \\
        20 & 3.2 & -19.6\% & 29.1 $\pm 0.2$ & 30.1 $\pm 0.1$ \\
        30 & 3.2 & -19.6\%  & 29.3 & 30.4 \\
        40 & 3.2 & -19.6\%  & 29.2 & 30.7 \\
    \bottomrule
    \end{tabular}
    \caption{\textbf{Intact vs. BoW on ImageNet-1k ZS.} Despite disadvantages like reduced dataset size and average caption length, we find that default BoW models result in a small but consistent improvement for both DeCLUTR and DistilBERT. The improvement is noticeably higher for DeCLUTR (+1.2 at 20 epochs). The improvement cannot be explained by randomness specific to a particular run since there is little variance between runs. Longer training also helps both intact captions and BoW captions.}
    \label{tab:deftxt-main}
\end{table}

\textbf{IN1k-ZS Results: } We compare intact captions against the BoW captions constructed with default deformation from Section \ref{sec:text-deformation} in Table \ref{tab:deftxt-main}. To recap, the following operations are applied to a random 90\% subset of captions: \texttt{Shuffle + RmStopNalpha + LimitToBaseVocab + RmTopFreq(1000) + Keep(4)}. Or, plainly, shuffle words, remove stop and non-alphabetical words, limit the words to a base vocabulary, remove any word that is in the set of \texttt{t = 1000} most frequent base vocabulary words, and keep \texttt{n = 4} words. It is worth noting that applying these operations leads to $\sim 20\%$ reduction in dataset size because some BoW are empty. Overall, it also reduces the average size of captions from 10.3 to 3.2. We observe consistent improvement in the BoW models despite 1) different architectures (DeCLUTER and DistilBERT) 2) longer training (20, 30, and 40 epochs) 3) different training seed (averaged across 4 different runs). Also, BoW models are data efficient, since they are better despite being trained on $\sim$ 20\% less data.

\textbf{Results on other datasets and architectures: }
In order to further support our observation that \emph{``BoW captions do not degrade Zero-Shot (ZS) performance significantly"} we repeat our experiments on 2 additional architectures and also do ZS evaluation on 7 additional datasets. The results are reported in Figure \ref{fig:def-vs-no-def-all-datasets}. We can see that BoW captions result in a significantly improved accuracy on Flowers102, DTD, and Pets for all architectures, but, it consistently degrades for Caltech101, Sun397 and Food101.

\begin{figure}
    \centering
    \includegraphics[width=0.45\columnwidth]{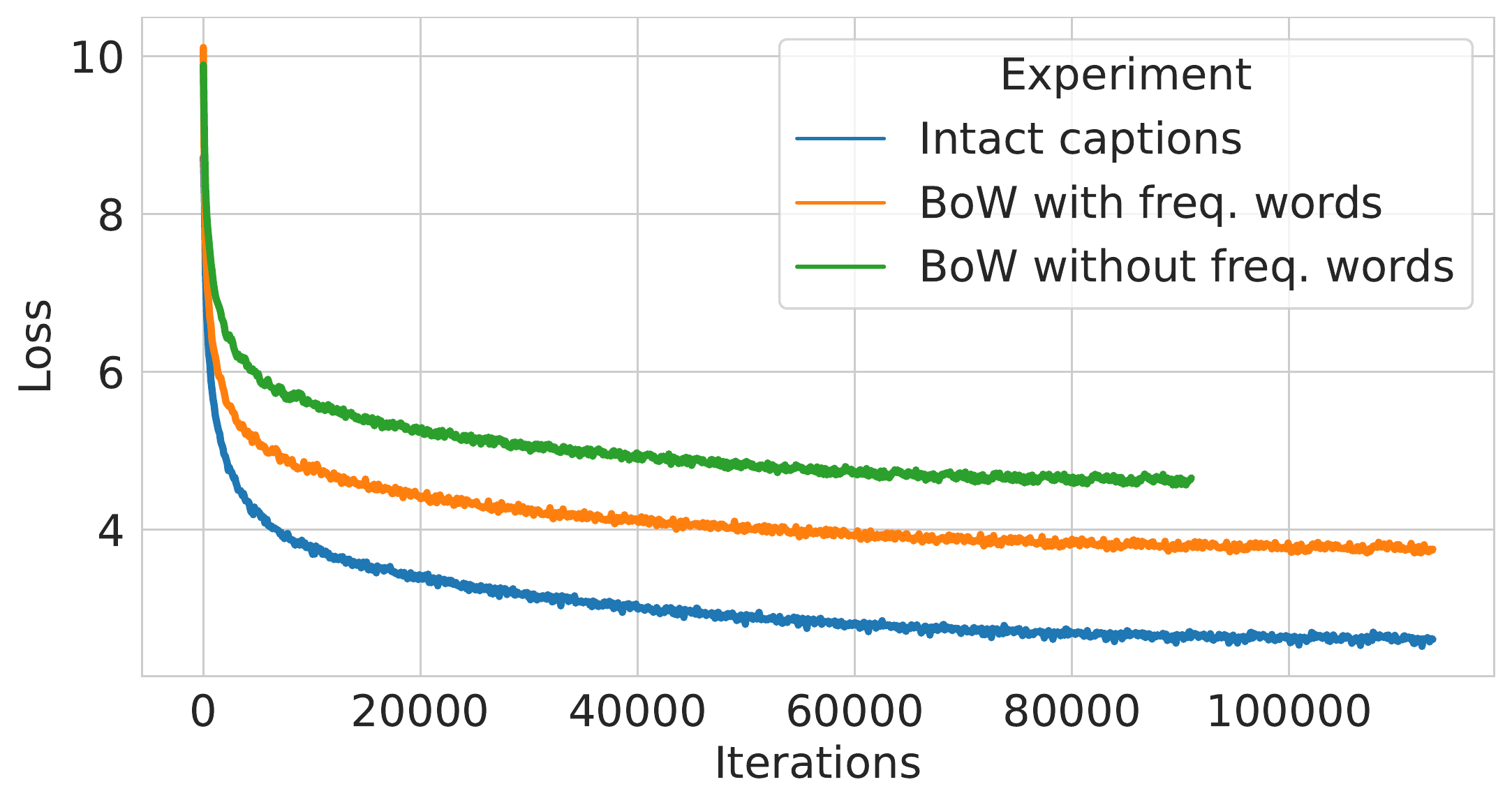}
    \includegraphics[width=0.45\columnwidth]{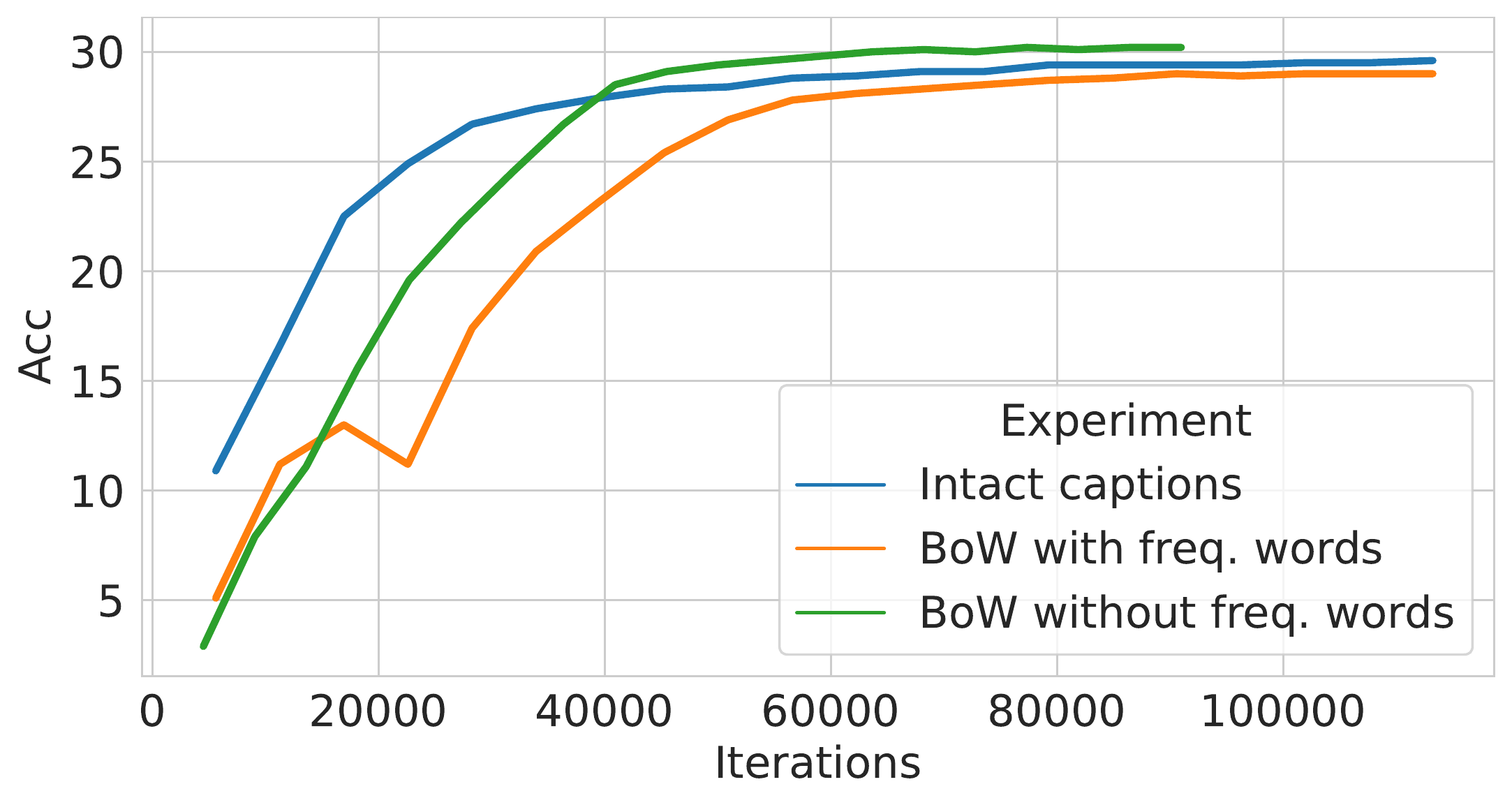}
    \caption{\textbf{Task hardness vs. BoW}. We hypothesize that removing frequent words makes the task harder since the model cannot rely on co-occurrence statistics of the frequent words to solve the task of distinguishing correct image-caption pairs during pre-training. The loss curves in the top figure shows evidence that supports our hypothesis. The curve with highest loss uses the default deformations to create BoW where 1k most frequent words are removed (BoW without frequent words). Unlike other deformations, it shifts the focus towards less-frequent concepts in the text. This is shown in the bottom figure where the IN-1k-ZS accuracy quickly converges to its final accuracy for intact captions, but it takes longer and reaches higher for the default deformation (BoW without frequent words). The deformation which keeps the frequent words (BoW with frequent words) simply makes the task harder but the model can still use frequent words to solve the task without learning other concepts.}
    \label{fig:loss-vs-iterations}
\end{figure}

\textbf{Why BoW captions help performance?:} We hypothesize that the default deformation which includes \texttt{RmTopFreq(1000)} operation removes words that make the contrastive task simpler when they are present. The model can solve contrastive task by simply latching on to the co-occurrence statistics of most frequent words while ignoring the rest. Thus, removing the most frequent words could make the contrastive task harder. Indeed, this does happen during training. Figure \ref{fig:loss-vs-iterations} compares the training loss curves of three DistilBERT experiments: intact captions, BoW with frequent words removed (default deformation), and BoW with frequent words kept (default deformation minus the \texttt{RmTopFreq(1000)} operation). We can see that the second one has the highest loss. Because the task is harder, we should also expect the model to take longer to reach its final ZS accuracy, and we indeed see this behavior in Figure \ref{fig:loss-vs-iterations}. But, despite the task being difficult for BoW with frequent words kept, the ZS accuracy does not improve beyond the intact captions baseline. Thus, simply increasing the difficulty of the pre-training task is not enough. The task also has to induce the model to learn less-frequent concepts. Note that any deformation will make the task harder since the text encoders are initialized with language models that have only seen sentences with correct syntax.

\toremove{
\textbf{Deformed vs Non-deformed on retrieval evaluation: } Despite being image-to-text retrieval at the core, zero-shot evaluation only tests the ability of a model to distinguish between a set of categories or nouns. This is easier compared to the task of cross-modal (image-to-text and text-to-image) retrieval on a dataset like MS-COCO \cite{mscoco} where the descriptions can be much more complex. Two image and text pairs might contain the same set of objects (nouns) but their relationship can be very different. Thus, it is reasonable to expect a model trained on deformed text descriptions to perform worse than its non-deformed counterpart in the zero-shot retrieval setting (Table \ref{tab:retrieval}). But, when both models are fine-tuned the gap between them shrinks greatly. This finding is consistent with U-VisualBERT \cite{u-visualbert} where a vision-and-language model without any paired data has comparable performance to the one that uses paired data for finetuning evaluation. Further, \cite{mlm_shuffle} shows that transformer based language models are robust to word order shuffle during pre-training for fine-tuning evaluation.

\begin{table}
    \centering
    \scalebox{0.85}{
    \begin{tabular}{@{}lcccccc@{}}
    \toprule
        & \multicolumn{6}{c}{MS-COCO} \\
        & \multicolumn{3}{c}{Image to Text} & \multicolumn{3}{c}{Text to Image}\\
        & R@1 & R@5 & R@10 & R@1 & R@5 & R@10 \\
        \midrule
        \multicolumn{7}{l}{\textit{Zero-Shot}} \\
        Visual N-Grams  & 8.7 & 23.1 & 33.3 & 5.0 & 14.5 & 21.9 \\
        Ours (defm)     & 17.7 & 39.5 & 52.0 & 13.7 & 33.2 & 45.5 \\
        Ours            & 23.0 & 48.2 & 62.2 & 19.0 & 42.0 & 53.8 \\
        CLIP            & 58.4 & 81.5 & 88.1 & 37.8 & 62.4 & 72.2 \\
        \midrule
        \multicolumn{7}{l}{\textit{Finetune}} \\
        Ours (defm) & 54.4 & 81.6 & 89.1 & 37.7 & 67.4 & 78.0 \\
        Ours        & 55.8 & 82.5 & 89.8 & 38.9 & 68.3 & 78.8 \\
    \bottomrule
    \end{tabular}
    }
    \caption{\textbf{Retrieval evaluation}. The task of caption retrieval is a harder task compared to zero-shot classification. As a result, the deformed models result in significantly worse performance compared to their non-deformed counterparts in zero-shot setting. But, when the models are finetuned the gap between them is greatly reduced. These findings are in-line with those of \cite{mlm_shuffle} which shows that language models are not affected by shuffled word order during pre-training. \wbc{Emphasize the DELTA. }}
    \label{tab:retrieval}
\end{table}
}

\begin{table}
    \centering
    \begin{tabular}{lccc}
    \toprule
        \multirow{3}{*}{Input} & \multicolumn{3}{c}{IN-1k-ZS} \\
         & \multicolumn{2}{c}{Freeze} & Train \\
         \cmidrule{2-3}
         & Text & Image & Both \\
         \midrule
        Intact captions & 15.1 & 26.7 & 29.5 \\
        BoW captions & 16.9 & 26.4 & 30.1 \\
    \bottomrule
    \end{tabular}
    \caption{\textbf{Effect of freezing the encoders.} In case of frozen text encoder, intact caption results are similar to BoW. This shows that despite being initialized from an already trained language model, presence or absence of syntax does not matter. Further, we find that freezing the image encoder hurts the model significantly less as compared to freezing the text model. This indicates that training text encoder is more important than the image encoder.}
    \label{tab:freeze-encoders}
\end{table}


\textbf{Ablations: } We explore freezing one of the encoders in Table \ref{tab:freeze-encoders}. We find that a frozen text encoder does not work better with intact captions than BoW captions despite being trained on natural language sentence during self-supervised training. Table \ref{tab:deftxt-ablation} shows the results for ablating different BoW construction techniques. The table shows that even when 100\% BoW captions are used the drop in accuracy is not significant. Moreover, keeping 10\% captions intact boosts the performance. We study the effect of removing different number of most frequent words from the base vocabulary in Table \ref{tab:deftxt-ablation-rmtop}. We also explore the effect of epochs, memory bank size, and text encoder architectures/initializations on an intact captions model in Table \ref{tab:baselines-ablations}. We can see that memory bank size = 8k and epochs = 20 work well for all architectures, and longer training only helps marginally. Thus, we chose these as the default parameters for all experiments. We chose DistilBert and DeCLUTR for other experiments since they are respectively the best and the worst.

\begin{table}
    \centering
    \scalebox{0.9}{
    \begin{tabular}{@{}lccc@{}}
    \toprule
        \multirow{3}{*}{Deformations} & Avg. & \multicolumn{2}{c}{IN1k-ZS} \\
        & Cap & De    & Distil \\
        & Len & CLUTR & BERT \\
        \midrule
        Intact captions & 10.3 & 27.9 & 29.5 \\
        \midrule
        \multicolumn{4}{@{}l}{\textit{only BoW}} \\
        \texttt{Shuffle} & 10.3 & 25.9 & 26.9 \\
        \quad \texttt{+ RmStopNalpha} & 5.8 & 27.8 & 28.5 \\
        \quad \quad \texttt{+ Keep(1)} & 1.0 & 16.7 & 18.3 \\
        \quad \quad \texttt{+ Keep(2)} & 2.0 & 24.2 & 26.1 \\
        \quad \quad \texttt{+ Keep(4)} & 3.8 & 27.3 & 27.9 \\
        \midrule
        \multicolumn{4}{@{}l}{\textit{mostly BoW}} \\
        \texttt{Shuffle} & 10.3 & 25.7 & 28.2 \\
        \quad \texttt{+ RmStopNalpha} & 6.3 & 28.2 & 28.9 \\
        \quad \quad \texttt{+ Keep(4)} & 4.5 & 27.8 & 29.1 \\
        \quad \quad \quad \texttt{+ LimitToBaseVocab} & 4.5 & 28.0 & 29.0 \\
        \quad \quad \quad \texttt{+ RmTopFreq(1000)} & 3.2 & 29.1 & 30.1 \\
    \bottomrule
    \end{tabular}
    }
    \caption{\textbf{Ablation of deformations: } Each subsequent row adds a new operation on top of a previous row with lesser indentation. ``mostly BoW" refers to the setting where 90\% of text is converted to BoW while in ``only BoW" all 100\% are BoW. Note that \texttt{Keep(n)} is always applied in the end after all other deformations.}
    \label{tab:deftxt-ablation}
\end{table}

\begin{table}
    \centering
    \scalebox{0.90}{
    \begin{tabular}{@{}lcccc@{}}
    \toprule
        & Avg. & $\Delta$ & \multicolumn{2}{c}{IN1k-ZS} \\
        & Cap & Dset & De    & Distil \\
        & Len & Size & CLUTR & BERT \\
        \midrule
        Intact captions & 10.3 & 0\% & 27.7 & 29.6 \\
        \midrule
        \texttt{RmTopFreq(0)} & 10.3 & 0\% & 27.8 & 29.1 \\
        \texttt{RmTopFreq(500)} & 3.4 & -10.3\% & 29.3 & 30.3 \\
        \texttt{RmTopFreq(1000)} & 3.2 & -19.6\% & 29.1 & 30.2 \\
        \texttt{RmTopFreq(2000)} & 3.1 & -34.1\% & 28.3 & 29.1 \\
        \texttt{RmTopFreq(4000)} & 3.4 & -53.6\% & 26.8 & 27.5 \\
        \texttt{RmTopFreq(8000)} & 4.7 & -73.5\% & 21.4 & 23.3 \\
    \bottomrule
    \end{tabular}
    }
    \caption{\textbf{Ablation for removing different number of most frequent words.} We explore different paramters for \texttt{RmTopFreq(t)} operation in the default deformation. We can see that removing too many frequent words reduces the dataset size which degrades the results. The average caption length increases for \texttt{t > 2000} since the overall dataset size decreases but the 10\% captions are always intact.}
    \label{tab:deftxt-ablation-rmtop}
\end{table}

\begin{table}
    \centering
    \scalebox{0.90}{
    \begin{tabular}{cccc}
    \toprule
        Memory & Epochs & \multicolumn{2}{c}{IN1k-ZS} \\
        \midrule
        & & DeCLUTR & DistilBERT \\
        \cmidrule[\lightrulewidth]{3-4}
        8k & 10 & 25.9 & 28.0 \\
        8k & 20 & 27.7 & 29.6 \\
        8k & 30 & 28.2 & 29.8 \\
        8k & 40 & 28.5 & 30.3 \\
        16k & 10 & 25.8 & 27.4 \\
        16k & 20 & 28.5 & 29.4 \\
        \midrule
        & & RoBERTa & BERT \\
        \cmidrule[\lightrulewidth]{3-4}
        8k & 10 & 25.3 & 26.5 \\
        8k & 20 & 28.2 & 28.5 \\
        16k & 10 & 25.2 & 25.3 \\
        16k & 20 & 27.3 & 28.0 \\
    \bottomrule
    \end{tabular}
    }
    \caption{\textbf{Ablation of training hyperparameters.} All models are trained with intact captions. We can see that 8k memory bank size and 20 epochs work well for all inits with DistilBERT being the best, and longer training only leads to marginal gains. Thus, we select 8k memory bank size and 20 epochs as the default setting for most experiments.}
    \label{tab:baselines-ablations}
\end{table}


\subsection{Semi-aligned Learning}
As we observed in previous section, the zero-shot performance of the model does not degrade by using BoW captions. Motivated by this observation we propose to generate BoW pseudo-labels for images that do not have any captions and add them to the training dataset. This setting is inspired by semi-supervised learning \cite{self-training} where unlabeled images are also included in training by pseudo-labeling them. In addition to the fully aligned CC dataset \cite{conceptual_captions} of size $\sim 3$M, we use 3M randomly sampled images from ImageNet-21k \cite{imagenet} during this experiment. Since only a part of the data is aligned, we call this setting ``semi-aligned''.

We use the DistilBERT model with 30.2\% on the IN-1k-ZS benchmark for pseudo-labeling. We selcted this model randomly from one of the 4 models trained with above configuration. It is a BoW model trained with default hyperparameters. The pseudo-labels are only calculated once at the beginning of the training after which they remain frozen. The results are reported in the Table \ref{tab:semi-aligned}. The total number of iterations used by a semi-aligned model are larger since it relies on an aligned-only model. Thus, we compare semi-aligned models with a longer trained baseline.

\begin{table}
    \centering
    \scalebox{0.97}{
    \begin{tabular}{@{}lcccc@{}}
    \toprule
        & Aligned & Un- & Epochs & IN1k \\
        &        & aligned & & ZS \\
        \midrule
        Aligned-only (ours) & 2.3M & - & 20 & 30.1 \\
        Aligned-only (ours) & 2.3M & - & 40 & 30.7 \\
        CLIP & 15M & - & 32 & 31.3 \\
        \midrule
        \texttt{RmTopRand} & 2.9M & 3M & 10 & 30.5 \\
        \texttt{RmTopMaxCount} & 2.9M & 3M & 10 & 31.1 \\
        \texttt{MaxCount} & 2.9M & 3M & 10 & 31.4 \\
        \texttt{WeightedCount} & 2.9M & 3M & 10 & 31.5 \\
    \bottomrule
    \end{tabular}
    }
    \caption{\textbf{Semi-aligned}. We explore using additional images without captions (unaligned). We experiment with various word ranking strategies for creating BoW pseudo caption and find that \texttt{WeightedCount} works the best and is on-par with the CLIP model that is trained with 15M aligned data. A model from the first row is used to for pseudo-labeling. We also compare with a longer trained baseline for a fair comparison.}
    \label{tab:semi-aligned}
\end{table}


\section{Related Works}


Our work is mainly related to vision and language pretraining, zero-shot learning, and text as supervision works which are described below, but our work also shares similarities with other areas. Our key idea of text deformation is also studied in \cite{mlm_shuffle} and \cite{unnatural} for natural language understanding tasks. Our proposed semi-aligned learning is inspired from the idea of self-training \cite{self-training} in semi-supervised learning literature \cite{semi-sup-orig}. Also, related is the work \cite{semi-sup-image-cap} which utilizes both paired and unpaired data for the task of image captioning. Finally, since our pre-training objective essentially optimizes cross-modal retrieval between images and text, our work is related to cross modal retrieval \cite{image-text-retrieval}.

\textbf{Vision and language pre-training (VLP):} Here, the goal is to learn generalizable cross-modal, vision and language, (V\&L) features during pre-training and then fine-tune them for complex downstream V\&L tasks. A BERT-style \cite{bert} transformer model is used to deeply fuse the representations from both modalities. VisualBERT \cite{visualbert} does early fusion while, ViLBERT \cite{vilbert} delays fusion until later layers. ALBEF \cite{albef} uses contrastive loss to improve alignment between features before fusing them. \cite{data_eff_mlm} improves the masked-language-modeling component of pre-training. \cite{12in1} 12-in-1 improves the finetuning stage by multi-task training on 12 tasks. U-VisualBERT \cite{u-visualbert} explores removing paired vision and language data during pre-training. \cite{vl-input-deform} explores the extent to which the model uses cross-modal cues by deforming individual modalities. See \cite{unmasked} for a more comprehensive overview of the words in this area. Unlike works in this area, we do not do complex modality fusion which requires parameter sharing between both modalities and focus on of zero-shot learning for evaluation.



\textbf{Zero-shot learning (ZSL): } Traditionally, zero-shot learning implies a model that can classify classes without seeing any examples (zero shot) during training \cite{zsl-review}. Initial works like DAP \cite{dap}, used semantic attribute classifiers, but later works like DeViSE began aligning the image and word features \cite{devise,zsl-socher}. See \cite{zsl-review} for a complete overview of this area. Our work is most similar to \cite{zsl-socher} where unsupervised image features are projected in the representation space of unsupervised word embeddings. Unlike traditional zero-shot learning, CLIP \cite{clip} defines it as generalization to novel image recognition datasets where it shows impressive performance by learning representations from scratch with a large-scale (400M) dataset of image and caption pairs. ALIGN \cite{align} further scales-up the pre-traininig dataset to noisier 1B data points. \cite{otter} proposes a distillation based loss to better handle the noise in the dataset. Efficient-CLIP \cite{efficientclip} is a concurrent work that uses text-only data to perform unimodal MLM task. DeCLIP \cite{declip} is another concurrent work that uses multiple unimodal SSL losses in addition to removing the noise from negatives with the help of nearest neighbors \cite{nn-contrastive,msf}. Our training and evaluation is similar to \cite{clip,declip,align}, but our goal is to shed more light on the supervision contained in aligned captions.

\textbf{Text as supervision: } Our work explores natural language sentences as a source of supervision. The work in \cite{joulin-bow} uses the objective of predicting the BoW tags associated with an image to train visual features. This work is extended in \cite{visual-ngrams} to predict n-grams instead of individual words. VirTex \cite{virtex} uses generative modeling of text as a data-efficient way of training visual features. CLIP \cite{clip} improves upon these works by using a contrastive loss that converges faster. We show that natural language sentences can be turned into BoW without hurting performance.

\section{Limitations \& Future Work}

Our work mainly focuses on understanding which components of the natural language supervision are important. Few directions which we could not explore in this work are as follows. What components of the text encoder architecture make it invariant to input deformations? While \cite{mlm_shuffle} shows that even large-scale pre-training is invariant to word order shuffle, we did not have the resources to verify it in our setting. Finally, limited gains from semi-aligned learning in Table \ref{tab:semi-aligned} indicate that it is a much harder task as compared to semi-supervised learning. All of the above limitations are important avenues for future works. Specifically, improving semi-aligned learning with different architectures and better pseudo-captioning techniques are important areas. 


\section{Conclusion}

We investigate what components of the natural language supervision in aligned image-caption pairs is truly required for zero-shot evaluation. We design various operators to deform captions and convert them into Bag-of-Words. We find that training models with BoW captions instead of natural language captions does not noticeably hurt the zero-shot classification performance. Furthermore, by carefully balancing the words in BoW we can even improve the performance. We apply this insight to design a recipe for utilizing unaligned images to further boost the visual concepts understanding. We use the model trained on aligned image-caption pairs to generate pseudo-captions BoW. We then use this new data alongside the original fully aligned data for pre-training a visual recognition model.\\

\vspace{-0.1in}
\noindent \textbf{Societal Impact:} Since we use self-supervised models trained on large-scale datasets for initialization, our models will inherit their biases. Specially, the text encoder is a large language model which has shown to learn the harmful biases against certain social groups.

{\small
\bibliographystyle{ieee_fullname}
\bibliography{egbib}
}

\newpage
\appendix
\section{Appendix}

\textbf{Implementation Details: }  We use PyTorch for implementation with SGD optimizer (weight\_decay = $0$, momentum = $0.9$), batch size = 512, and a cosine learning rate decay with initial LR = $0.003$. The model is trained on 8 A100 GPUs which requires 18hrs for 20 epochs of full 2.9M CC dataset. The temperature is set to $0.02$, and the default memory bank size is 8192 ($\sim$ 8k). We use a linear layer on top of the vision encoder to match the dimension of its embeddings with the text embeddings. Since the dimension of text encoder embeddings is typically 768, the linear layer is $2048\times768$ since we ResNet-50 as the vision encoder. For the creating a single sentence level embedding from a list of tokens, we use average pooling. We also perform an ablation over using \texttt{[CLS]} pooling but we did not find any noticeably difference in performance. The comparison between average and \texttt{[CLS]} pooling methods can be found in Table \ref{tab:avg-vs-cls-pooling}.

\begin{table}
    \centering
    \scalebox{0.85}{
    \begin{tabular}{@{}lcccccc@{}}
    \toprule
        & \multicolumn{6}{c}{MS-COCO} \\
        & \multicolumn{3}{c}{Image to Text} & \multicolumn{3}{c}{Text to Image}\\
        & R@1 & R@5 & R@10 & R@1 & R@5 & R@10 \\
        \midrule
        \multicolumn{7}{l}{\textit{Zero-Shot}} \\
        Visual N-Grams  & 8.7 & 23.1 & 33.3 & 5.0 & 14.5 & 21.9 \\
        Ours (BoW)     & 17.7 & 39.5 & 52.0 & 13.7 & 33.2 & 45.5 \\
        Ours (Intact)  & 23.0 & 48.2 & 62.2 & 19.0 & 42.0 & 53.8 \\
        CLIP            & 58.4 & 81.5 & 88.1 & 37.8 & 62.4 & 72.2 \\
        \midrule
        \multicolumn{7}{l}{\textit{Finetune}} \\
        Ours (BoW)      & 54.4 & 81.6 & 89.1 & 37.7 & 67.4 & 78.0 \\
        Ours (Intact)   & 55.8 & 82.5 & 89.8 & 38.9 & 68.3 & 78.8 \\
    \bottomrule
    \end{tabular}
    }
    \caption{\textbf{BoW vs Intact captions on retrieval evaluation}. The task of caption retrieval is a harder task compared to zero-shot classification. As a result, the BoW model results in a significantly worse performance compared to its intact captions counterpart in the zero-shot setting. But, when the models are fine-tuned the gap between them is greatly reduced. These findings are similar to those of \cite{mlm_shuffle,u-visualbert} which show that input deformations during pre-training do not hurt the models when they are fine-tuned for downstream tasks. For Image to text retrieval with R@1, we see that the gap between Ours (BoW) and Ours (Intact) is about 6 points but when both of them are fine-tuned the gap reduces to about 1 point.}
    \label{tab:retrieval}
\end{table}

\textbf{BoW vs intact captions on retrieval evaluation: } Despite being image-to-text retrieval at the core, zero-shot evaluation only tests the ability of a model to distinguish between a set of categories or nouns. This is easier compared to the task of cross-modal (image-to-text and text-to-image) retrieval on a dataset like MS-COCO \cite{mscoco} where the descriptions can be much more complex. Thus, it is reasonable to expect a model trained on BoW text descriptions to perform worse than its intact counterpart in the zero-shot retrieval setting. We indeed see this behaviour in Table \ref{tab:retrieval} but, we also see that when both models are fine-tuned the gap between them shrinks greatly. This finding is consistent with U-VisualBERT \cite{u-visualbert} where a vision-and-language model without any paired data has comparable performance to the one that uses paired data for fine-tuning evaluation. Further, \cite{mlm_shuffle} shows that transformer based language models are robust to word order shuffle during pre-training for fine-tuning evaluation.

\begin{table}
    \centering
    \scalebox{0.8}{
    \begin{tabular}{llcc}
    \toprule
        & & \multicolumn{2}{c}{IN-1k-ZS} \\
        & & Avg-pooling & \texttt{[CLS]} \\
        \midrule
        \multirow{2}{*}{RoBERTa}    & Intact    & 28.3 & 28.3 \\
                                    & BoW       & 28.7 & 28.9 \\
                                    & $\Delta$  & +0.4 & +0.6 \\
        \midrule
        \multirow{2}{*}{BERT}   & Intact    & 28.4 & 28.1 \\
                                & BoW       & 28.2 & 28.4 \\
                                & $\Delta$  & -0.2 & +0.3 \\
        \midrule
        \multirow{2}{*}{DeCLUTR}   & Intact    & 27.9 & 27.9 \\
                                & BoW       & 29.1  & 28.9 \\
                                & $\Delta$  & +1.2 & +1.0 \\
        \midrule
        \multirow{2}{*}{DistilBERT} & Intact    & 29.5 & 29.3 \\
                                    & BoW       & 30.1 & 29.8 \\
                                    & $\Delta$  & +0.6 & +0.5 \\
    \bottomrule
    \end{tabular}
    }
    \caption{\textbf{Average vs. \texttt{[CLS]} pooling:} We explore the effect of using \texttt{[CLS]} pooling strategy to go from a list of tokens to a sentence level embedding. Averarge pooling is the default strategy, but we show that switching to \texttt{[CLS]} does not significantly change the results. The BoW models are still better than their intact counterparts by similar margins.}
    \label{tab:avg-vs-cls-pooling}
\end{table}

\textbf{Transfer linear evaluation:} Similar to \cite{clip,byol,simclr}, we also evaluate the transferability of our trained models by training linear layers on top of frozen vision encoders. The training and evaluation procedure is exactly the same as MSF \cite{msf}, and the results are reported in Table \ref{tab:linear-evaluation}.


\begin{figure*}
    \centering
    \includegraphics[width=\textwidth]{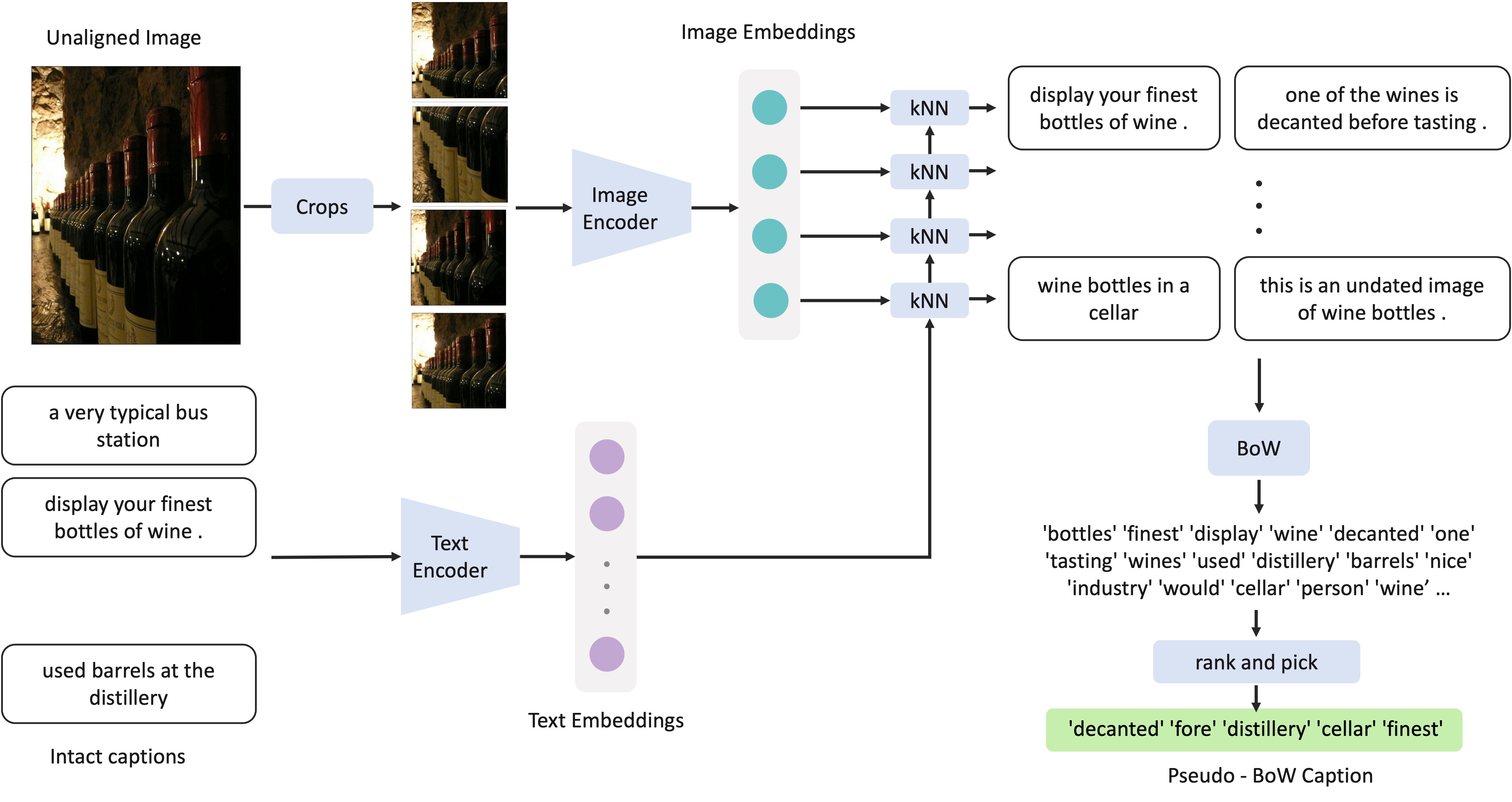}
    \caption{\textbf{Generating pseudo BoW captions:} We illustrate the process of constructing pseudo BoW captions for unaligned images in this figure. A fully aligned dataset, CC \cite{conceptual_captions}, to train the vision and text encoders and to obtain the intact captions used for retrieval. We find nearest neighbors in the text domain for different crops of a given unaligned image. The retrieved captions of all crops are then aggregated into a BoW. The words are then ranked according to one of the strategies mentioned in Section 2.3 of main text, and a few top words (4 in this figure) are chosen as the pseudo BoW caption. We used the \texttt{WeightedCount} strategy in this figure.}
    \label{fig:pseudo-bow-generation}
\end{figure*}

\begin{table*}
    \centering
    \scalebox{0.8}{
    \begin{tabular}{ll|llllllll|l}
    \toprule
        & & ImageNet & Food101 & CIFAR100 & Sun397 & DTD & Pets & Caltech101 & Flowers & Mean \\
        \midrule
        \multirow{2}{*}{RoBERTa}    & Intact    & 28.3 (\small{$\pm 0.4$)} & 17.8 (\small{$\pm 0.3$)} & 31.1 (\small{$\pm 0.9$)} & 38.6 (\small{$\pm 0.2$)} & 23.2 (\small{$\pm 1.1$)} & 17.5 (\small{$\pm 0.7$)} & 71.7 (\small{$\pm 1.7$)} & 12.8 (\small{$\pm 0.6$)} & 30.1 \\
                                    & BoW       & 28.7 (\small{$\pm 0.8$)} & 15.3 (\small{$\pm 0.7$)} & 32.8 (\small{$\pm 2.7$)} & 35.0 (\small{$\pm 0.7$)} & 25.6 (\small{$\pm 1.0$)} & 22.8 (\small{$\pm 0.8$)} & 71.2 (\small{$\pm 1.2$)} & 14.2 (\small{$\pm 0.9$)} & 30.7 \\
                                    & $\Delta$  & +0.4 & -2.5 & +1.7 & -3.5 & +2.4 & +5.3 & -0.5 & +1.4 & +0.6 \\
        \midrule
        \multirow{2}{*}{BERT}   & Intact    & 28.4 (\small{$\pm 0.2$)} & 18.0 (\small{$\pm 0.4$)} & 30.9 (\small{$\pm 1.4$)} & 35.1 (\small{$\pm 0.5$)} & 26.0 (\small{$\pm 0.6$)} & 20.4 (\small{$\pm 0.8$)} & 70.5 (\small{$\pm 1.9$)} & 16.1 (\small{$\pm 0.8$)} & 30.7 \\
                                & BoW       & 28.2 (\small{$\pm 0.1$)} & 17.2 (\small{$\pm 0.8$)} & 30.8 (\small{$\pm 2.0$)} & 32.4 (\small{$\pm 0.2$)} & 30.5 (\small{$\pm 1.1$)} & 26.5 (\small{$\pm 1.0$)} & 66.8 (\small{$\pm 1.2$)} & 16.6 (\small{$\pm 0.7$)} & 31.1 \\
                                & $\Delta$  & -0.2 & -0.8 & -0.0 & -2.7 & +4.5 & +6.0 & -3.8 & +0.5 & +0.4 \\
        \midrule
        \multirow{2}{*}{DeCLUTR}   & Intact    & 27.9 (\small{$\pm 0.2$)} & 17.6 (\small{$\pm 0.6$)} & 28.5 (\small{$\pm 1.1$)} & 36.0 (\small{$\pm 0.3$)} & 24.3 (\small{$\pm 1.2$)} & 19.7 (\small{$\pm 0.8$)} & 69.3 (\small{$\pm 1.3$)} & 15.5 (\small{$\pm 0.8$)} & 29.9 \\
                                & BoW       & 29.1 (\small{$\pm 0.2$)} & 17.2 (\small{$\pm 0.5$)} & 32.3 (\small{$\pm 1.7$)} & 33.8 (\small{$\pm 0.2$)} & 31.7 (\small{$\pm 1.1$)} & 24.9 (\small{$\pm 0.4$)} & 68.4 (\small{$\pm 3.0$)} & 19.1 (\small{$\pm 0.7$)} & 32.1 \\
                                & $\Delta$  & +1.2 & -0.4 & +3.8 & -2.1 & +7.4 & +5.2 & -0.9 & +3.5 & +2.2 \\
        \midrule
        \multirow{2}{*}{DistilBERT} & Intact    & 29.5 (\small{$\pm 0.1$)} & 17.2 (\small{$\pm 0.4$)} & 30.8 (\small{$\pm 1.4$)} & 38.0 (\small{$\pm 0.4$)} & 23.6 (\small{$\pm 1.3$)} & 20.8 (\small{$\pm 1.0$)} & 72.3 (\small{$\pm 0.7$)} & 13.5 (\small{$\pm 0.6$)} & 30.7 \\
                                    & BoW       & 30.1 (\small{$\pm 0.1$)} & 17.1 (\small{$\pm 0.5$)} & 33.7 (\small{$\pm 1.7$)} & 36.3 (\small{$\pm 0.2$)} & 29.9 (\small{$\pm 1.0$)} & 25.9 (\small{$\pm 1.0$)} & 72.0 (\small{$\pm 0.6$)} & 15.6 (\small{$\pm 0.6$)} & 32.6 \\
                                    & $\Delta$  & +0.6 & -0.1 & +2.9 & -1.8 & +6.3 & +5.0 & -0.2 & +2.1 & +1.9 \\
    \bottomrule
    \end{tabular}
    }
    \caption{\textbf{Detailed zero-shot results:} We report detailed zero-shot results here. These results were used to generate the Figure 3 of the main text. We can see that the results are consistently improved for all architectures on Pets by minimum $\sim 5$ points and on DTD by minimum $\sim 2$ points. On average, BoW is better than using intact captions by $0.6$ points for RoBERTa, $0.4$ points for BERT, $2.2$ points for DeCLUTR, and $1.9$ points for DistilBERT. This shows that the model's improvements with BoW are relatively higher compared to using intact captions.}
    \label{tab:detailed-zero-shot}
\end{table*}

\begin{table*}
    \centering
    \scalebox{0.8}{
    \begin{tabular}{ll|llllllll}
    \toprule
        & & Food101 & CIFAR100 & Sun397 & DTD & Pets & Caltech101 & Flowers & Mean \\
        \midrule
        BYOL \cite{byol} & - & 75.3 & 78.4 & 62.2 & 75.5 & 90.4 & 94.2 & 96.1 & 81.7 \\
        \midrule
        \multirow{2}{*}{RoBERTa} & Intact  & 77.7 (\small{$\pm0.2$}) & 78.6 (\small{$\pm0.4$}) & 66.4 (\small{$\pm0.1$}) & 76.8 (\small{$\pm0.2$}) & 90.6 (\small{$\pm0.1$}) & 94.9 (\small{$\pm0.1$}) & 96.5 (\small{$\pm0.1$}) & 83.1 \\
                                 & BoW      & 77.5 (\small{$\pm0.0$}) & 78.2 (\small{$\pm0.1$}) & 65.6 (\small{$\pm0.1$}) & 76.0 (\small{$\pm0.3$}) & 91.0 (\small{$\pm0.3$}) & 94.7 (\small{$\pm0.2$}) & 96.3 (\small{$\pm0.1$}) & 82.8 \\
        \midrule
        \multirow{2}{*}{BERT}   & Intact    & 77.7 (\small{$\pm0.3$}) & 79.1 (\small{$\pm0.3$}) & 66.1 (\small{$\pm0.1$}) & 76.7 (\small{$\pm0.2$}) & 90.6 (\small{$\pm0.2$}) & 94.8 (\small{$\pm0.0$}) & 96.7 (\small{$\pm0.1$}) & 83.1 \\
                                & BoW       & 77.8 (\small{$\pm0.2$}) & 78.6 (\small{$\pm0.4$}) & 65.5 (\small{$\pm0.0$}) & 76.3 (\small{$\pm0.2$}) & 91.1 (\small{$\pm0.1$}) & 94.8 (\small{$\pm0.1$}) & 96.4 (\small{$\pm0.1$}) & 82.9 \\
        \midrule
        \multirow{2}{*}{DeCLUTR}& Intact    & 77.9 (\small{$\pm0.1$}) & 78.9 (\small{$\pm0.3$}) & 66.3 (\small{$\pm0.1$}) & 77.0 (\small{$\pm0.2$}) & 90.7 (\small{$\pm0.1$}) & 94.8 (\small{$\pm0.1$}) & 96.4 (\small{$\pm0.1$}) & 83.1 \\
                                & BoW       & 77.7 (\small{$\pm0.1$}) & 78.5 (\small{$\pm0.2$}) & 65.7 (\small{$\pm0.1$}) & 76.2 (\small{$\pm0.3$}) & 90.9 (\small{$\pm0.1$}) & 94.6 (\small{$\pm0.2$}) & 96.4 (\small{$\pm0.0$}) & 82.8 \\
        \midrule
        \multirow{2}{*}{DistilBERT} & Intact & 77.8 (\small{$\pm0.2$}) & 78.8 (\small{$\pm0.3$}) & 66.5 (\small{$\pm0.1$}) & 76.3 (\small{$\pm0.3$}) & 90.3 (\small{$\pm0.8$}) & 94.8 (\small{$\pm0.1$}) & 96.5 (\small{$\pm0.2$}) & 83.0 \\
                                    & BoW    & 77.6 (\small{$\pm0.2$}) & 78.2 (\small{$\pm0.3$}) & 65.7 (\small{$\pm0.1$}) & 75.8 (\small{$\pm0.3$}) & 90.9 (\small{$\pm0.3$}) & 94.7 (\small{$\pm0.2$}) & 96.5 (\small{$\pm0.1$}) & 82.8 \\
    \bottomrule
    \end{tabular}
    }
    \caption{\textbf{Transfer linear evaluation:} We evaluate the models by training linear probes over frozen vision encoders. We see little difference between using BoW and intact caption models. We also see little effect of the text encoder's architecture over the performance of the vision encoder. Note that unlike the zero-shot setting, only the vision encoder is used during evaluation. Finally, we see that both BoW and intact caption models are better than the BYOL \cite{byol} initialization. This suggests that natural language supervision improves the quality of the visual features.}
    \label{tab:linear-evaluation}
\end{table*}

\begin{figure*}
    \centering
    \includegraphics[width=\textwidth]{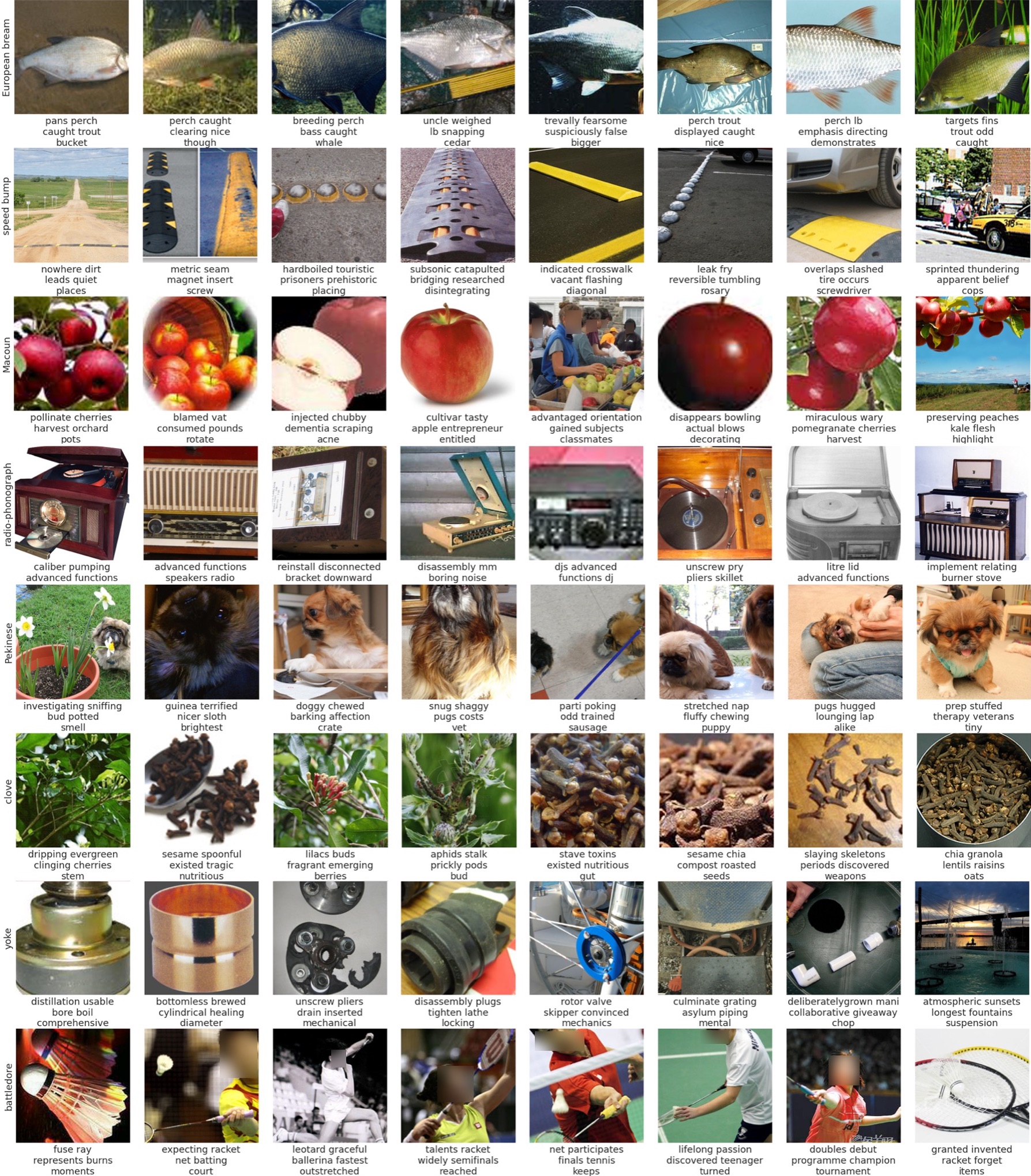}
    \caption{\textbf{Examples of pseudo BoW captions:} We show some examples of the pseudo BoW captions generated by the \texttt{WeightedCount} strategy. Each row contains images from a single category. The actual category name is listed on the left, at the start of each row, and the pseudo captions are below each image.}
    \label{fig:unaligned-1}
\end{figure*}


\end{document}